\documentclass[letterpaper, 10 pt, conference]{ieeeconf}  

\IEEEoverridecommandlockouts                              

\usepackage{graphics} 
\usepackage{graphicx} 
\usepackage{epsfig} 
\usepackage{amsmath, bm} 
\usepackage{amssymb}  
\usepackage{physics}
\usepackage{upgreek}
\usepackage{multicol}
\usepackage{multirow}
\usepackage{lipsum}  
\usepackage{breqn}
\usepackage{comment} 
\usepackage{booktabs}
\usepackage[linesnumbered, ruled]{algorithm2e}

\PassOptionsToPackage{hyphens}{url}

\usepackage{xcolor}

\usepackage[style=ieee]{biblatex}

\DeclareSourcemap{
  \maps{
    \map{
      \pertype{article}
      \step[fieldset=language, null]
      \step[fieldset=url, null]
      \step[fieldset=doi, null]
      \step[fieldset=issn, null]
      \step[fieldset=isbn, null]
      \step[fieldset=note, null]
      \step[fieldset=editor, null]
      \step[fieldset=urldate, null]
      \step[fieldset=file, null]
    }
  }
}
\DeclareSourcemap{
  \maps{
    \map{
      \pertype{inproceedings}
      \step[fieldset=language, null]
      \step[fieldset=url, null]
      \step[fieldset=doi, null]
      \step[fieldset=issn, null]
      \step[fieldset=isbn, null]
      \step[fieldset=note, null]
      \step[fieldset=editor, null]
      \step[fieldset=urldate, null]
      \step[fieldset=file, null]
    }
  }
}
\DeclareSourcemap{
  \maps{
    \map{
      \pertype{incollection}
      \step[fieldset=language, null]
      \step[fieldset=url, null]
      \step[fieldset=doi, null]
      \step[fieldset=issn, null]
      \step[fieldset=isbn, null]
      \step[fieldset=note, null]
      \step[fieldset=editor, null]
      \step[fieldset=urldate, null]
      \step[fieldset=file, null]
    }
  }
}

\title{\LARGE \bf
Efficient Path Planning and Tracking for Multi-Modal Legged-Aerial Locomotion Using Integrated Probabilistic Road Maps (PRM) and Reference Governors (RG)   
}

\author{Eric Sihite$^{1}$, Benjamin Mottis$^{2}$, Paul Ghanem$^{3}$, Alireza Ramezani$^{3}$, and Morteza Gharib$^{1}$%
\thanks{$^{1}$ The author is with the Department of Aerospace, California Institute of Technology, Pasadena, CA-91125, USA. (e-mail: esihite, mgharib@caltech.edu).}%
\thanks{$^{2}$ The author is with the Department of Microtechnology, Ecole Polytechnique Fédérale de Lausanne, 1015 Lausanne, Switzerland. He is doing his Master’s Thesis with the Department of Aerospace, California Institute of Technology, Pasadena, CA-91125, USA. (e-mail: benjamin.mottis@epfl.ch).}%
\thanks{$^{3}$ The author is with the SiliconSynapse Laboratory, Department of Electrical and Computer Engineering, Northeastern University, Boston, MA-02119, USA. (e-mail: ghanem.p, a.ramezani@northeastern.edu)}%
}

\begin{document}

\maketitle
\thispagestyle{empty}
\pagestyle{empty}

\begin{abstract}

There have been several successful implementations of bio-inspired legged robots that can trot, walk, and hop robustly even in the presence of significant unplanned disturbances. Despite all of these accomplishments, practical control and high-level decision-making algorithms in multi-modal legged systems are overlooked. In nature, animals such as birds impressively showcase multiple modes of mobility including legged and aerial locomotion. They are capable of performing robust locomotion over large walls, tight spaces, and can recover from unpredictable situations such as sudden gusts or slippery surfaces. Inspired by these animals' versatility and ability to combine legged and aerial mobility to negotiate their environment, our main goal is to design and control legged robots that integrate two completely different forms of locomotion, ground and aerial mobility, in a single platform. Our robot, the \textit{Husky Carbon}, is being developed to integrate aerial and legged locomotion and to transform between legged and aerial mobility. This work utilizes a Reference Governor (RG) based on low-level control of Husky's dynamical model to maintain the efficiency of legged locomotion, uses Probabilistic Road Maps (PRM) and 3D A$^\star$ algorithms to generate an optimal path based on the energetic cost of transport for legged and aerial mobility. 

\end{abstract}


\section{Introduction}
\label{sec:introduction}


Raibert's hopping robots \cite{raibert1984experiments}, and Boston Dynamic's BigDog \cite{raibert2008bigdog} are amongst the most successful examples of legged robots, as they can hop or trot robustly even in the presence of significant unplanned disturbances. Other than these successful examples, many bipedal and anthropomorphic robots have also been introduced \cite{ramezani_atrias_2012,ramezani_performance_2014,buss_preliminary_2014,park_finite-state_2013,park_switching_2012, dangol_towards_2020,dangol2021hzd,de2020thruster}. Boston Dynamics' dynamic humanoid, ATLAS, has pushed the limits of dynamic legged locomotion with its 28 hydraulically actuated joints. This robot has showcased impressive mobility feats, including jumping over obstacles and dynamic flip-turns. 

Despite all of these accomplishments, state-of-the-art legged robots are prone to fall-over and cannot negotiate highly rough terrains when they face large obstacles. In nature, animals such as birds impressively showcase multiple modes of mobility including legged and aerial locomotion. Birds and other animals are known for their natural, efficient, and robust locomotion feats and can fly over larger walls, inside tight spaces, can recover from unpredictable situations such as severe external pushes, scuffing, or slippage on icy surfaces. 

\begin{figure}
    \centering
    \vspace{0.1in}
    \includegraphics[width = 0.8\linewidth]{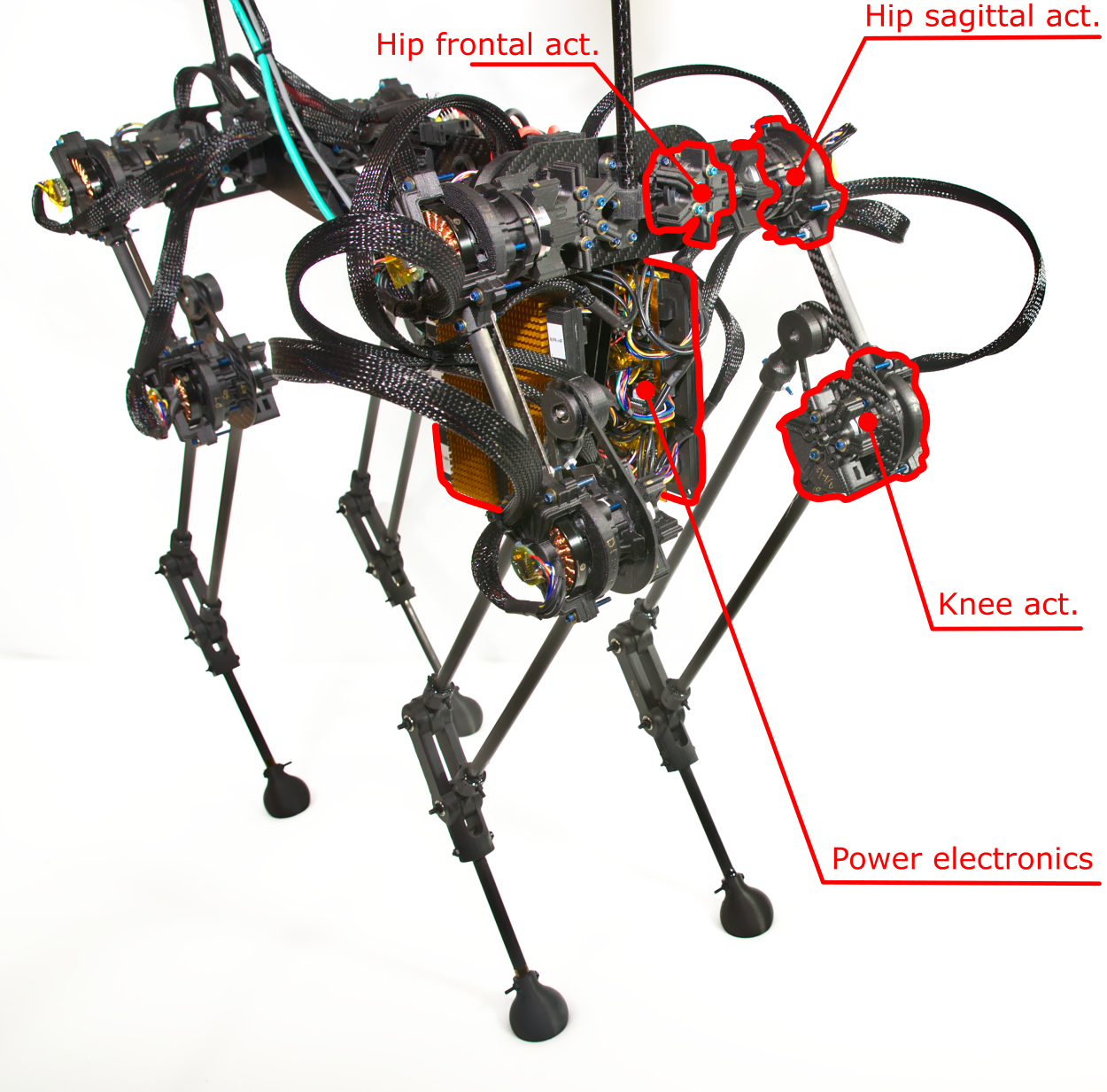}
    \caption{Illustrates NU's Husky Carbon Platform designed to explored multi-modal mobility in unstructured spaces \cite{ramezani_generative_2021}.}
    \vspace{-0.1in}
    \label{fig:cover}
\end{figure}

Inspired by animals multi-modal mobility, our main goal is to design and control legged robots that integrate two completely different forms of locomotion in a single platform. This paper will report our recent efforts in dynamic modeling and designing closed-loop feedback for the thruster-assisted locomotion of a quadrupedal legged robot called Northeastern University (NU) Husky Carbon, which is shown in Fig.~\ref{fig:cover}. Currently, Husky Carbon's hardware is being developed at NU. We have reported the successful legged locomotion of Husky in \cite{sihite2021optimization, ramezani_generative_2021, sihite2021unilateral}. The flight tests and integration of legged and aerial mobility are ongoing at the time the report is being prepared. First, we briefly discuss the previous work done on the path planning of multi-modal robot and present a rough overview of Husky's hardware.

\subsection{Multi-Modal Path Planning Past Work}

In order to take full advantage of the multi-modal capacities of the Husky, it is necessary to develop a path planning optimization methods that can incorporate Husky's multi-modal locomotion capability. Numerous researches have already been done on multi-modal robots which are either able to roll and fly such that the HyFDR \cite{sharif_energy_2018}\cite{sharif2019new} and the Drivocopter \cite{suh_optimal_2019} or to drive and navigate on water such that the Ambot \cite{suh_optimal_2019}. Most of the methods developed in these articles use a uniform discretization of the space, and then the optimal path is found with the Dijkstra algorithm \cite{suh_optimal_2019}, or with the A$^{\star}$ \cite{sharif_energy_2018}\cite{araki_multi-robot_2017}. Furthermore, in \cite{suh_optimal_2019}, an optimization technique based on a reduced model of the system is used to calculate the costs of the edges and then to smoother the final trajectory. Araki et al. \cite{araki_multi-robot_2017} have coupled their path planning method to a prioritization algorithm allowing swarm operation with 20 flying cars. While in this article \cite{sharif2019new}, Sharif et al. have developed an algorithm to select the locomotion mode of the HyFDR robot allowing to optimize the transport cost during outdoor navigation with only a 2D map of the environment.

\subsection{Overview of Husky Carbon Platform}

The design of Husky Carbon intends to achieve both quadrupedal mobility and multi-rotor flight within the same mechanical architecture. To this end, a propeller motor is attached to the outside of each knee joint, allowing the robot to morph into a quad-rotor configuration by extension of the hip frontal joints. There are three actuated degrees of freedom per leg: hip frontal flexion/extension, hip sagittal flexion/extension, and knee flexion/extension. To simplify the design for this initial prototype, off-shelf servomotors are used to actuate each joint in lieu of lighter, more specialized custom hardware. Extensive use of carbon fiber epoxy laminates fortify the airframe and leg bones, while 3D printed components with carbon fiber reinforcement serve as connecting members. The electronics are mounted on two vertical carbon fiber plates to yield a minimized Total Cost of Transport (TCoT) and payload \cite{ramezani_generative_2021}.

\section{Reduced-Order Model (ROM) Formulation}
\label{sec:dynamics}

\begin{figure}
    \centering
    \vspace{0.1in}
    \includegraphics[width = \linewidth]{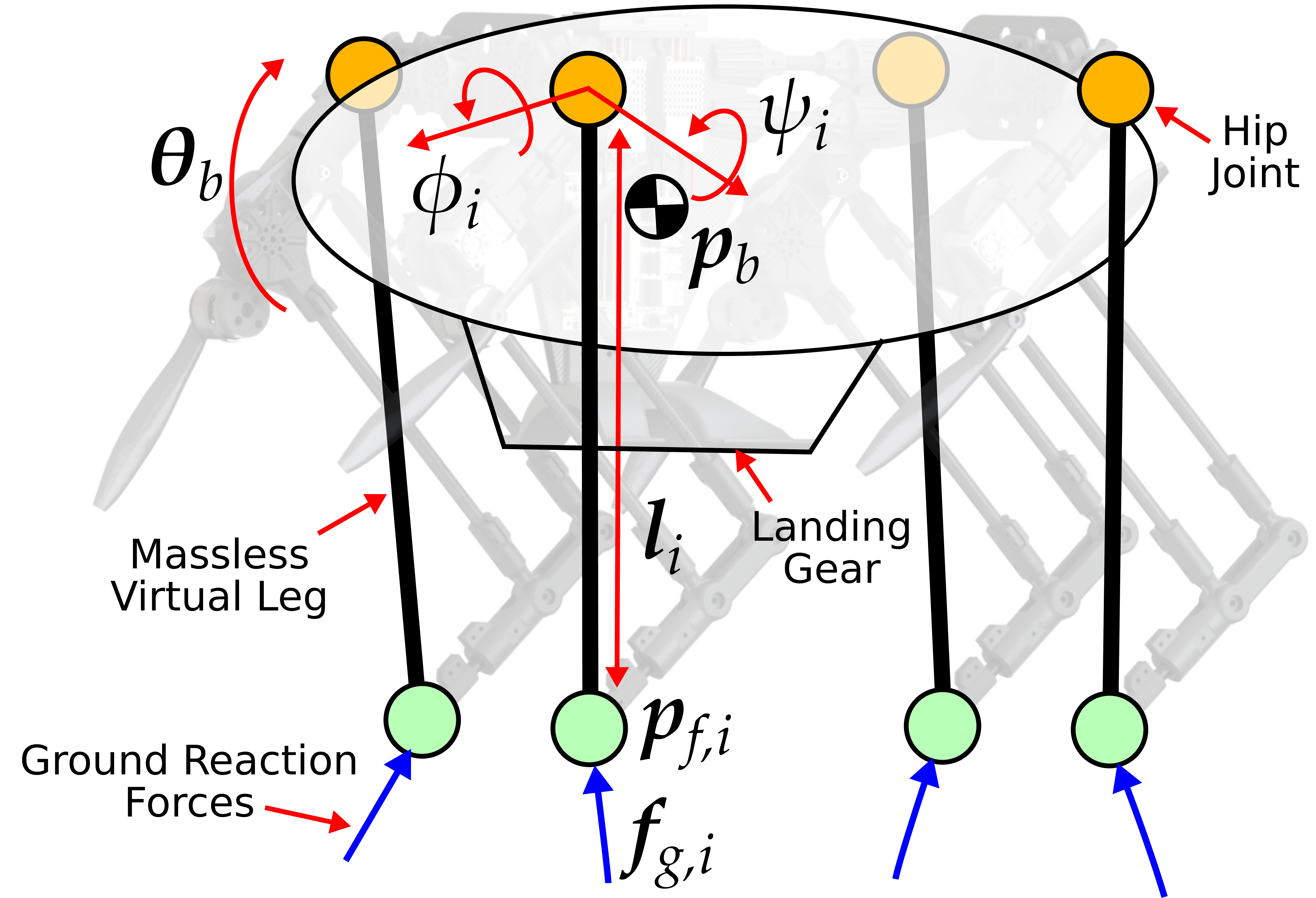}
    \caption{Illustrates the reduced-order model of Husky used for locomotion control and cost calculations in the A$^\star$ path search algorithm. This model simplifies the robot by assuming massless legs, which significantly reduces the complexity to a 6 DOF dynamical model. The thruster forces are applied at a fixed position along the leg, aligned with the hip sagittal axis.}
    \vspace{-0.1in}
    \label{fig:fbd}
\end{figure}

We developed a simulator using a reduced-order model (ROM) to simplify the trajectory tracking and cost calculations in the path search algorithm. This ROM assumes massless leg linkages and can be reduced down to a single body, 6 DOF dynamics. In this simplified model, each leg has 3 DOF to describe the foot position. These 3 DOF of leg $i = {1,2,3,4}$ are the hip frontal angle ($\phi_i$), hip sagittal angles ($\psi_i$), and leg length ($l_i$), as illustrated in Fig. \ref{fig:fbd}. This results in a total of 12 kinematics DOF and 6 dynamical DOF which is much simpler than the full dynamical model of the robot. The dynamical model can be derived using Euler-Lagrangian formulation. 

Let $\bm q_d = [\bm p_b^\top, \bm \theta_b^\top]^\top \in \mathbb{R}^6$ be the dynamical states, where $\bm p_b \in \mathbb{R}^3$ is the body center of mass (COM) inertial position and $\bm \theta_b \in \mathbb{R}^3$ is the Euler angles for the transformation from body frame to the inertial frame. Let $\bm q_k \in \mathbb{R}^{12}$ be the kinematic states of the virtual legs. The equation of motion for the dynamical system can simply be represented in the following form:
\begin{equation}
    M(\bm q_d)\,  \ddot{\bm q}_d + \bm h(\bm q_d, \dot{\bm q}_d) = \sum_{i=1}^4 \bm u_{g,i} + \sum_{i=1}^4 \bm u_{t,i},,
\end{equation}
where $M \in \mathbb{R}^{6 \times 6}$ is the inertial matrix, $\bm h \in \mathbb{R}^{6}$ contains the gravitational and coriolis terms, $\bm u_{g,i}$ and $\bm u_{t,i}$ is the generalized ground reaction forces (GRF) and thruster forces of leg $i \in \{1,2,3,4\}$, respectively. 

The forces acting on the dynamical body can be derived using virtual displacement to map the forces into the generalized coordinates $\bm q_d$. Let $\bm p_{f,i} \in \mathbb{R}^{3}$ and $\bm p_{t,i} \in \mathbb{R}^{3}$ be the foot and thrusters inertial positions of leg $i$. The generalized forces of both the GRF and thrusters can be derived as follows:
\begin{equation}
\begin{aligned}
    \bm u_{g,i} &= \left( \partial \dot{\bm p}_{f,i} /  \partial \dot{\bm q}_d  \right)^\top \, \bm f_{g,i} \\
    \bm u_{t,i} &= \left( \partial \dot{\bm p}_{t,i} /  \partial \dot{\bm q}_d  \right)^\top \, \bm f_{t,i},,
\end{aligned}
\end{equation}
where $\bm f_{g,i} \in \mathbb{R}^3$ and $\bm f_{t,i} \in \mathbb{R}^3$ are the GRF and thruster force defined in the inertial frame. The GRF $\bm f_{g,i}$ can be derived using a compliant ground model and Stribeck friction model for the forces normal and along the ground surface, respectively. Assuming flat ground surface, let $u_{g,x}$, $u_{g,y}$, and $u_{g,z}$ be the inertial force components of $\bm u_{g,i}$. The GRF can be defined as follows:
\begin{equation}
\begin{aligned}
    u_{g,z} =& -k_{g,p}\, p_{f,z} - k_{g,d}\, \dot p_{f,z} \\
    u_{g,x} =& -\left(\mu_c + (\mu_s - \mu_c)\, e^{-( |\dot p_{f,x}|^2\,/\, v_s^2)} \right) u_{g,z}\, \mathrm{sgn}(\dot p_{f,x}) \\ 
    & - \mu_v\,\dot p_{f,x}\\
\end{aligned}
\label{eq:ground_model}
\end{equation}
where $\mu_c$, $\mu_s$, and $\mu_v$ are the dry, static, and viscous friction coefficients, respectively. The friction along the $y-$axis ($u_{g,y}$) follows a similar derivations to $u_{g,x}$. Finally, the thruster force and torque can be defined as force and torque acting parallel to the hip sagittal axis.

\section{Low-level Locomotion Control, High-level Decision Making and Path Planning}
\label{sec:path_planning}

\subsection{Reference-Governor (RG) Based Control of Legged Locomotion}

Here, we assume a conventional flight control design which is skipped for brevity of this report. However, the optimality of the low-level legged locomotion control in terms of achieving feasible gaits is enforced within an RG-based framework. The RG framework is utilized to enforce the friction pyramid constraint in \eqref{eq:ground_model} by manipulating the applied reference into the kinematic states $\bm q_k$ \cite{dangol2020performance, liang2021rough,sihite_integrated_2021}. This method is very useful as it avoids using optimization frameworks to enforce locomotion feasibility constraints which as a result facilitates faster high-level decision making.

\begin{figure}
    \centering
    \vspace{0.1in}
    \includegraphics[width = 0.8\linewidth]{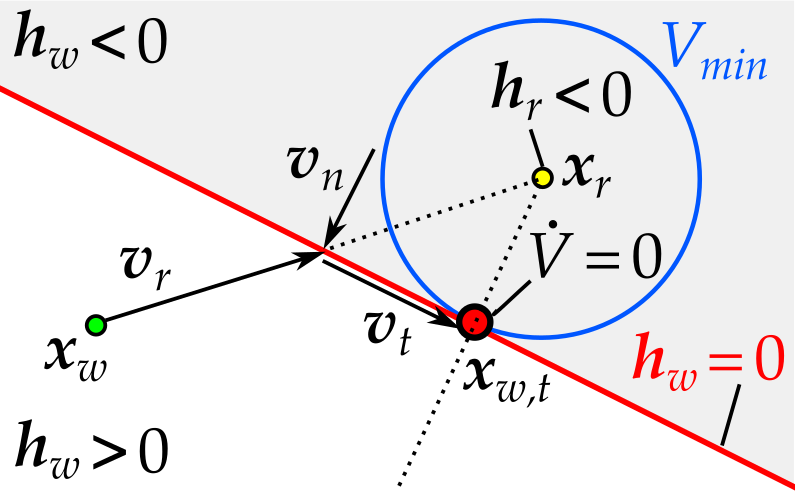}
    \caption{The Reference Governor update law to enforce ground friction constraint. The update directions $\bm v_r$, $\bm v_t$, and $\bm v_n$ directs the applied control reference $\bm x_w$ to the minimum energy solution $\bm x_{w,t}$ that is the closest to the desired reference $\bm x_r$ without breaking the constraint $\bm h_w = 0$.}
    \vspace{-0.1in}
    \label{fig:erg}
\end{figure}

Let $\bm x_w$ be the applied reference to $\bm q_k$ which will be used instead of the pre-defined (nominal) references $\bm x_r$. Also, consider the GRF constraints as a nonlinear function of $\bm x_w$ and ROM states denoted by $\bm h_w = \bm h_r(\bm q_d, \dot{\bm q}_d, \bm x_w)$. The RG algorithm manipulates the applied reference ($\bm x_w$) to avoid violating the constraint equation $\bm h_w \geq 0$ while also be as close as possible to the desired reference ($\bm x_r$), as illustrated in Fig.~\ref{fig:erg}. Consider the Lyapunov equation $V = (\bm x_r - \bm x_w)^\top P (\bm x_r - \bm x_w)$; $\bm x_w$ is updated through the update law:
\begin{equation}
    \dot{\bm x}_w = \bm v_r + \bm v_t + \bm v_n,
\label{eq:erg_update}
\end{equation}
where $\bm v_r$ drives $\bm x_w$ directly to $\bm x_r$, while $\bm v_t$ and $\bm v_n$ drives $\bm x_w$ along the surface and into the boundary $\bm h_w =  0$, respectively. The objective of this RG algorithm is to drive $\bm x_w$ to the state $\bm x_{w,t}$ which is the minimum energy solution $V_{min}$ that satisfies the constraint $\bm h_w \geq 0$. We denote the rowspace of the violated constraints of $\bm{h}_r$ by $C_r$. We define $N_r = \mathrm{null}(C_r) = [\bm{n}_1, \dots, \bm{n}_{n}]$ where $n$ is the size of the nullspace. Then the following update law is used for the term in \eqref{eq:erg_update}
\begin{equation}
\begin{aligned}
    \bm v_r &= \hat \alpha_r\, (\bm x_r - \bm x_w), \\
    \bm v_n &= \hat \alpha_n\,\bm r_k\,\bm r_k^\top\, (\bm x_r - \bm x_w) \\
    \bm v_t &= \textstyle \sum^n_{k=1} \hat{\alpha}_t\, \bm{n}_k\,\bm{n}_k^\top (\bm{x}_r - \bm{x}_w)
\end{aligned}
\label{eq:erg_update_v}
\end{equation}
where $\hat \alpha$ are scalars defined as follows:
\begin{equation}
\begin{aligned}
    \hat \alpha_r &= 
    \begin{cases}
     \alpha_r, & \text{if } \min(\bm h_w) \geq 0 \text{ or } \min(\bm h_r) \geq 0 \\
     0,      & \text{else} \\
    \end{cases} \\
    \hat \alpha_t &= 
    \begin{cases}
     \alpha_t, & \text{if } \min(\bm h_w) \geq 0 \text{ or } \min(\bm h_r) < 0 \\
     0,      & \text{else} \\
    \end{cases} \\
    \hat \alpha_n &= 
    \begin{cases}
     \alpha_n, & \text{if } \min(\bm h_w) \leq \min(\bm h_r) < 0 \\
     -\alpha_n, & \text{if } \min(\bm h_r) < \min(\bm h_w) < 0 \\
     0,      & \text{else,} \\
    \end{cases}
\end{aligned}
\label{eq:erg_update_alpha}
\end{equation}
where $\alpha_i$ is a positive scalar which determines the rate of convergence. 

\begin{figure}
    \centering
    \vspace{0.1in}
    \includegraphics[width = 0.8\linewidth]{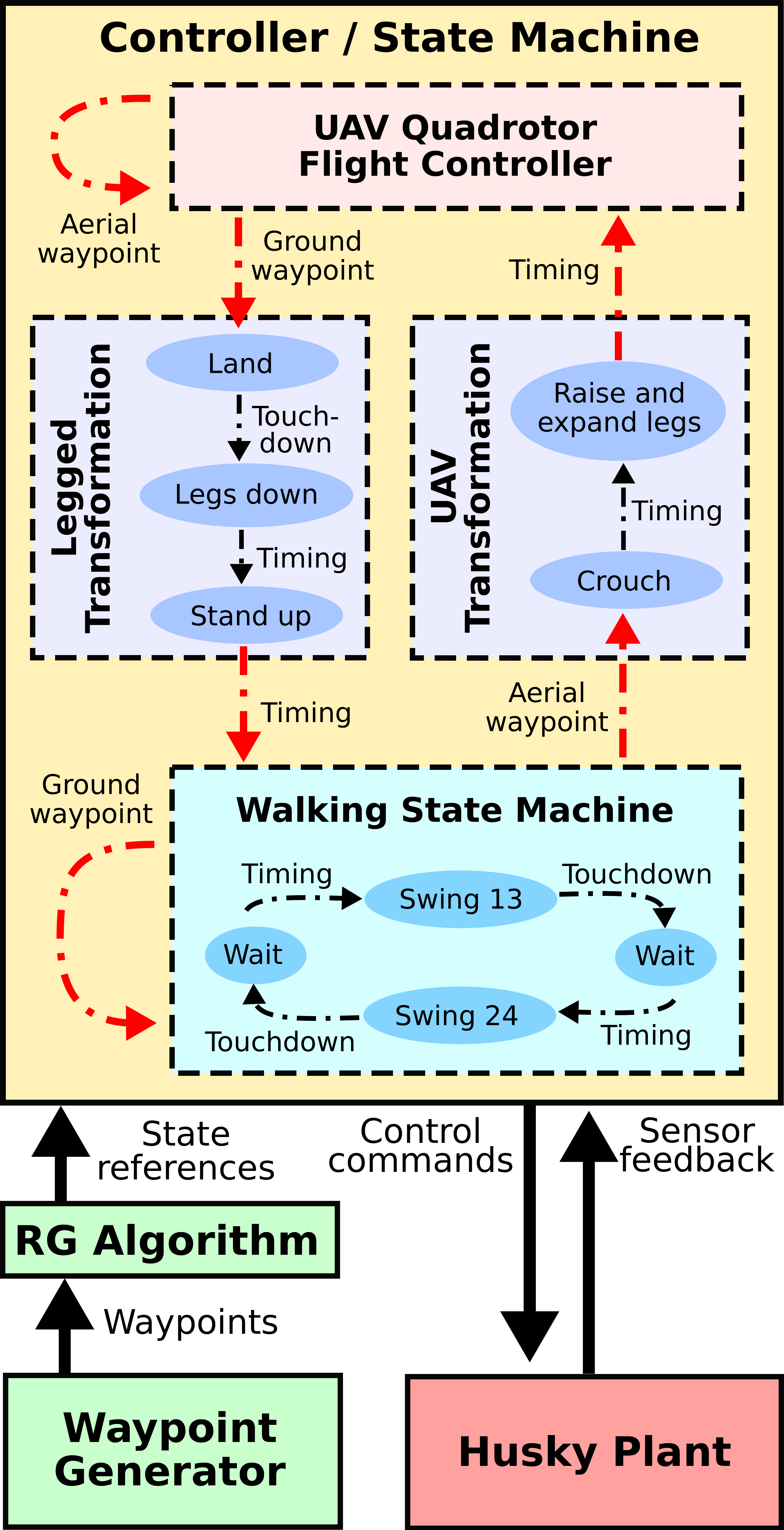}
    \caption{Low-level Locomotion Control Architecture and High-level Decision Making State Machine. The dashed arrow lines represent the switching surfaces of the state machine.}
    \vspace{-0.1in}
    \label{fig:control_architecture}
\end{figure}

The robot follows the waypoints generated by the path planning algorithm using a simple state machine showed in Fig. \ref{fig:control_architecture}. This state machine allows the robot to transform between the legged and aerial mobility by executing the transformation sequence whenever the waypoint switches the mode of locomotion (e.g., from legged to aerial, or vice versa). Then, the state machine provides the state references for the joints and flight controller to track.

The ground mobility controller follows a simple turning and forward walking speed tracking controller which are used in a similar fashion to a unicycle model. Given a waypoint, the robot will turn to face the target waypoint and walk forward until it reaches the destination. The aerial mobility controller follows a typical quadrotor flight controller scheme using two pairs of clockwise and counter-clockwise rotating propellers to generate the yaw moment and thrusts.

The transformation sequence follows a set routine done within a fixed time and rate. Transforming from legged to aerial mobility starts by raising the legs vertically upwards relative to the body, which effectively crouches the robot until the landing gear touches the floor, then followed by the hip joints rotation and leg length adjustment to the UAV configuration. The reverse of this sequence is used to transform the robot back to the legged mobility.

\subsection{High-level Decision Making and Path Planning}

The objective of the path planning strategy is to minimize the total energy consumed by the robot by optimizing the choice of the locomotion mode. To achieve this goal, the environment is first discretized into a set of nodes each associated with a locomotion mode (walking or flying). The nodes are then connected by edges, and a cost is computed for each of them. Finally, an A$^\star$algorithm is used to determine the optimal path defined by a set of waypoints, each associated with a state (Flying and Walking).

\subsubsection{Discretization Of The Environment}

Two different discretization methods have been used to create a set of nodes and edges representing the environment, and their performances are then compared. The first one consists in dividing the space into a set of uniformly distributed points. While, in the second one, the 3D environment is discretized into a set of nodes and edges with the 3D MM-PRM shown in Algorithm \ref{alg:3D_PRM}. Like in \cite{MM_PRM}, this adapted version of the Probabilistic Road Map (PRM) algorithm takes into account the Multi-Modal nature of the robot's movements.

The classical PRM algorithm builds a graph in the defined space by generating a certain number of nodes, where the nodes are created with random position one by one. When a node is created, it will search for the nearest nodes already present in the graph and then connect to them to form edges while checking that it does not cross any obstacles.  This method is adapted to generate a graph for unimodal robots by constraining the node generation to a single mode (e.g., create only ground nodes for a legged robot or create nodes in aerial space for a quadcopter). 

In this work, Husky can move both on the ground and in aerial space. Therefore, it is necessary to create 2 sets of constraints when generating the nodes. Thus, the main difference with the classical PRM algorithm is that a constraint is added on a certain number of nodes to ensure a sufficient number of nodes in each mode. This extended version of the PRM algorithm requires the definition of 3 parameters: the number of ground surface nodes $N_w$, the number of nodes describing flyable space $N_f$, and the maximum distance between neighboring nodes $R$.




%

New ground nodes $X_{new}$ are randomly assigned according to the following constraint:
\begin{equation}
X_{new} \in \{(x, y,z): z=z_{GND})\}.
\end{equation}
\noindent Similarly, new nodes in the flyable task space are obtained as follows:
\begin{equation}
X_{new} \in \{ (x, y, z): z > 0, z \ne z_{GND} \}.
\end{equation}

\begin{algorithm}[t]

\caption{3D MM-PRM Algorithm}\label{alg:3D_PRM}
\KwIn{$R$ radius of neighbors, $N_w$ number of walking node, $N_f$  number of flying nodes}
\KwOut{$N$ and $E$ respectively sets of nodes and edges}
$N \gets \emptyset$\;
$E \gets \emptyset$\;
\While{$n \leq (N_w + N_f)$} {
    \uIf{$n \leq N_w$}{
    $Xnew \gets random\_walking\_node()$\;
    }
    \Else{
    $Xnew \gets random\_flying\_node()$\;
    }
    \If{$obstacles\_free(Xnew)$}{
        $N \cup Xnew$\;
        $n \gets n + 1$\;
        $Xnearest \gets nearest(N, R, Xnew)$\;
        \For{$node \in Xnearest$}{
            \If {$clear\_edge(Xnew, node)$}{
            $E \cup \{Xnew, node\}$\;
            }
        } 
    }
}
\Return{$N,E$}
\end{algorithm}

\begin{figure}[t]
    \centering
    \vspace{0.1in}
    \includegraphics[width = \linewidth]{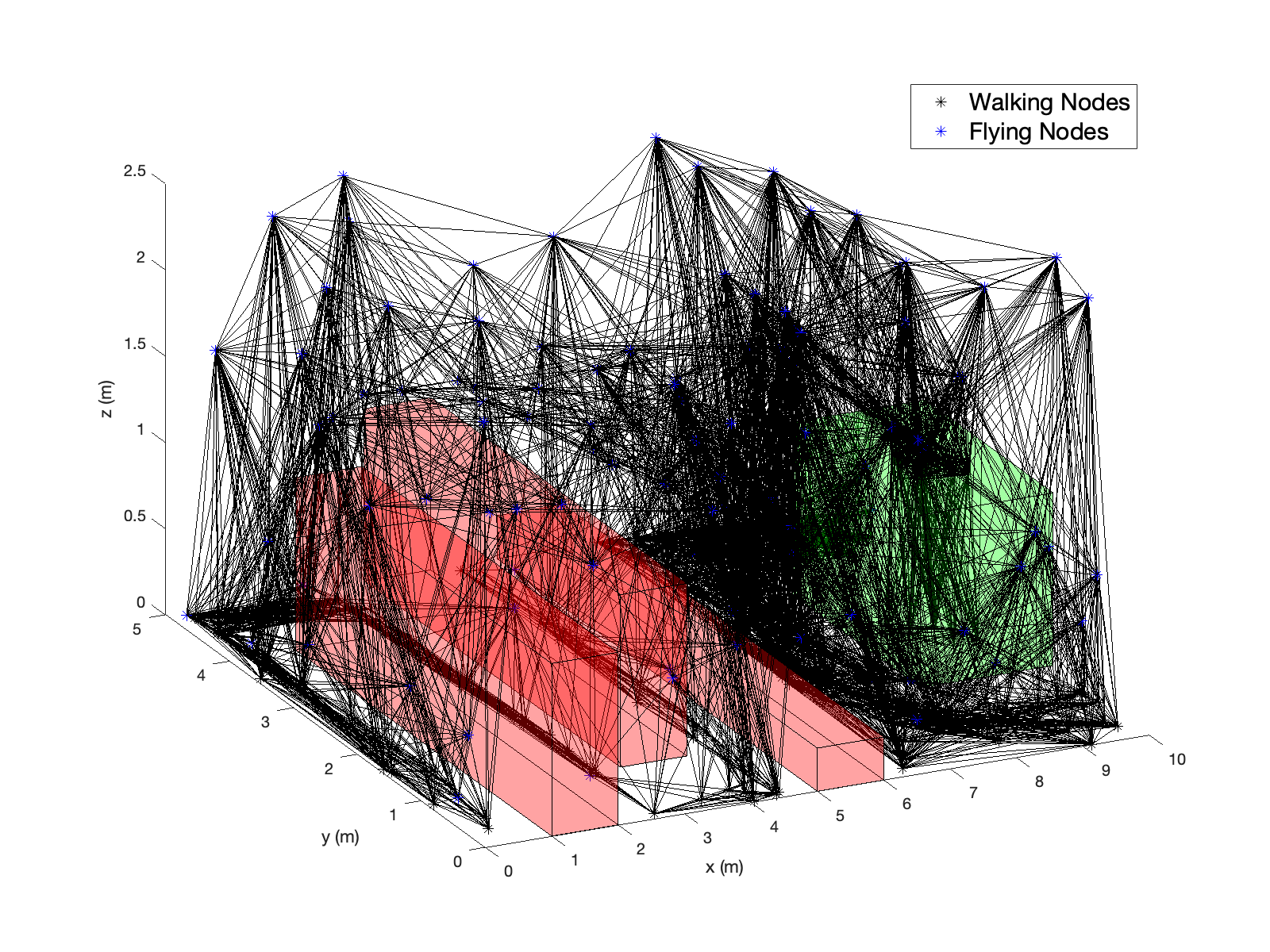}
    \caption{Example of graph generated by the 3D MM-PRM Algorithm with the following parameters: $R = 4$ meters, $N_w = 300$, and $N_f = 300$.}
    \vspace{-0.1in}
    \label{fig:ex_PRM}
\end{figure}

\begin{figure}[t]
    \centering
    \vspace{0.1in}
    \includegraphics[width = 0.8\linewidth]{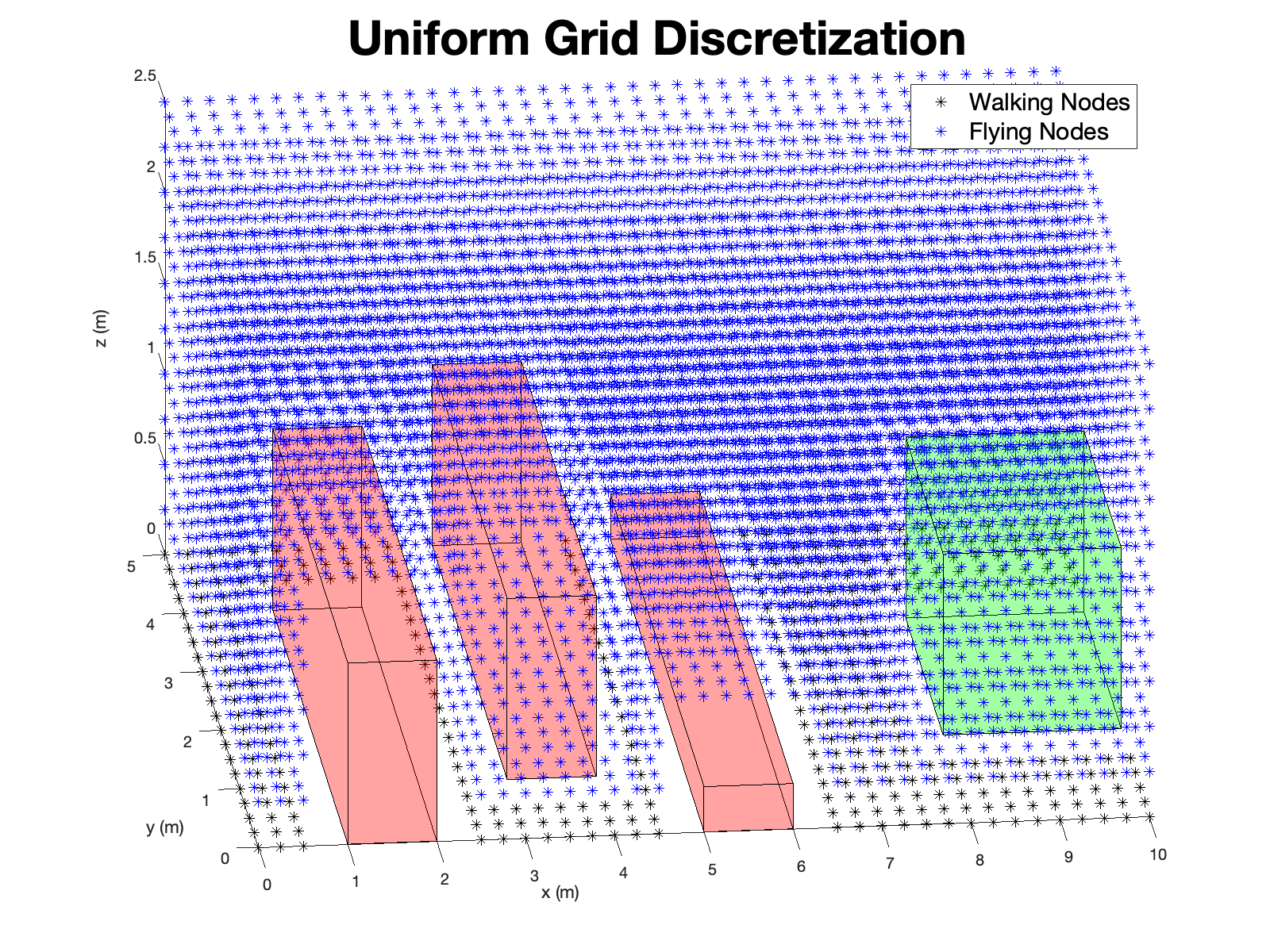}
    \includegraphics[width = 0.8\linewidth]{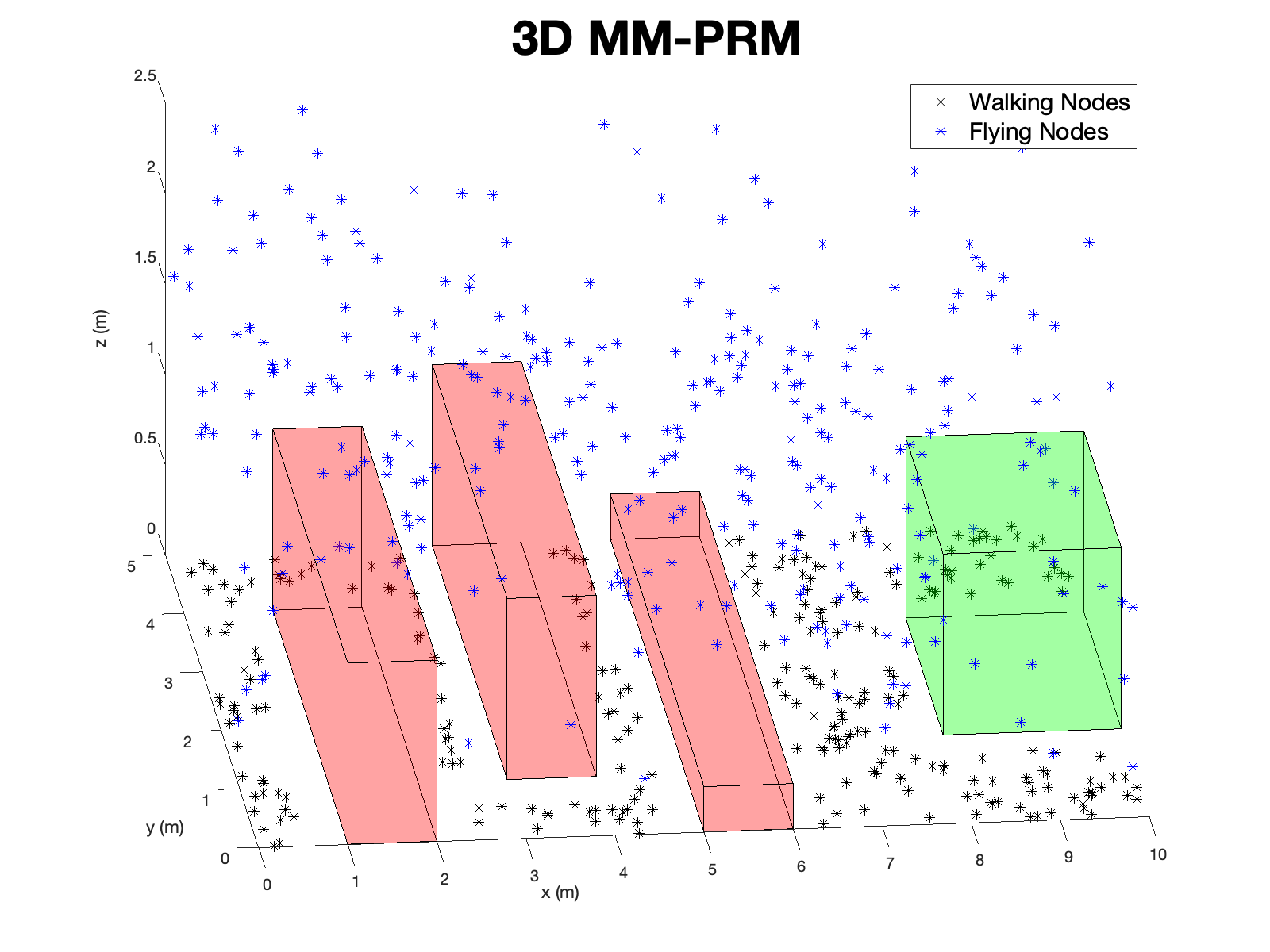}
    \caption{Representation of the set of nodes generated by the two discretization methods. The MM-RPM method generates a significantly reduced amount of nodes which greatly reduces the computational time and cost in performing the path finding algorithm.}
    \vspace{-0.1in}
    \label{fig:node_comparison}
\end{figure}


The search for neighboring nodes that will then be used to create the edges ($E$) is at the core of the PRM algorithm and is found using the following condition:
\begin{equation}
    X_{Nearest} = \{X \in \mathcal{N} : \norm{X_{new} - X} \leq R\},
\end{equation}
\noindent where $\mathcal{N}$ is the set of nodes already created, $R$ denotes the maximum radius distance, and $\norm{.}$ is the Euclidean norm. 

The cost and time of calculation are very strongly linked to the choice of the values of the algorithm parameters ($R$, $N_w$, $N_f$). The greater the total number of nodes or the greater the radius of acceptance of the neighbors, the greater the computation time and cost will be. Therefore, it is necessary to study the convergence of the result in function of the parameters in order to optimize to computation cost. We identified the parameters that led to best results. The parameters are $R = 4$ meters, $N_w = 300$ and $N_f = 300$. An example of the graph built with the 3D MM-PRM algorithm is presented in the Fig.~\ref{fig:ex_PRM}.

We found that compared to a uniform discretization with 0.25m-wide grids, the 3D MM-PRM algorithm produces a graph representative of the environment with a minimal number of nodes as shown in Fig.~\ref{fig:node_comparison}. This reduces the cost and the computing time while avoiding any compromises on the performance concerning the optimality of the path obtained. The comparison between these two methods is summarized briefly in Table~\ref{tab:comp_2_methods}, which shows the significant reduction in computational time when using the PRM algorithm.

\begin{table}[t]
\caption{Comparison of the two discretization methods}
\label{tab:comp_2_methods}
\centering
\begin{tabular}{|l|r|r|}
\hline
                         & \multicolumn{1}{c|}{3D MM-PRM} & \multicolumn{1}{c|}{Uniform Grid} \\ \hline
Number of Nodes              & 500                      & 9892                         \\ 
Number of Edges             & 30920                    & 219340                       \\ 
Computation Time {[}s{]} & 12.1                     & 78.29                        \\ \hline
\end{tabular}

\end{table}

\subsubsection{Calculation of Cost of Locomotion}
\label{sec:cost_calculation}

\begin{figure*}[t]
    \centering
    \vspace{0.1in}
    \includegraphics[width = 0.30\linewidth]{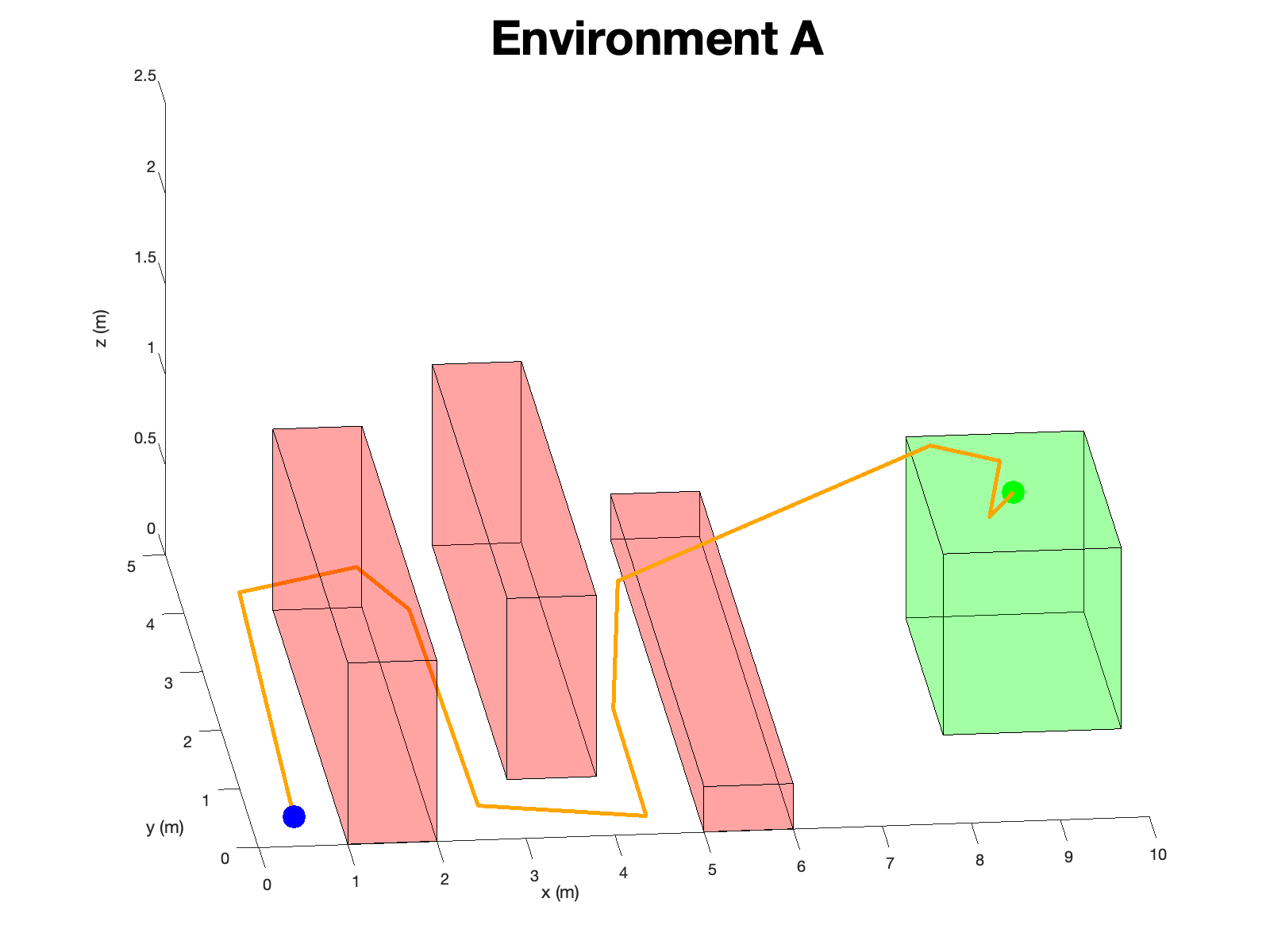}
    \includegraphics[width = 0.30\linewidth]{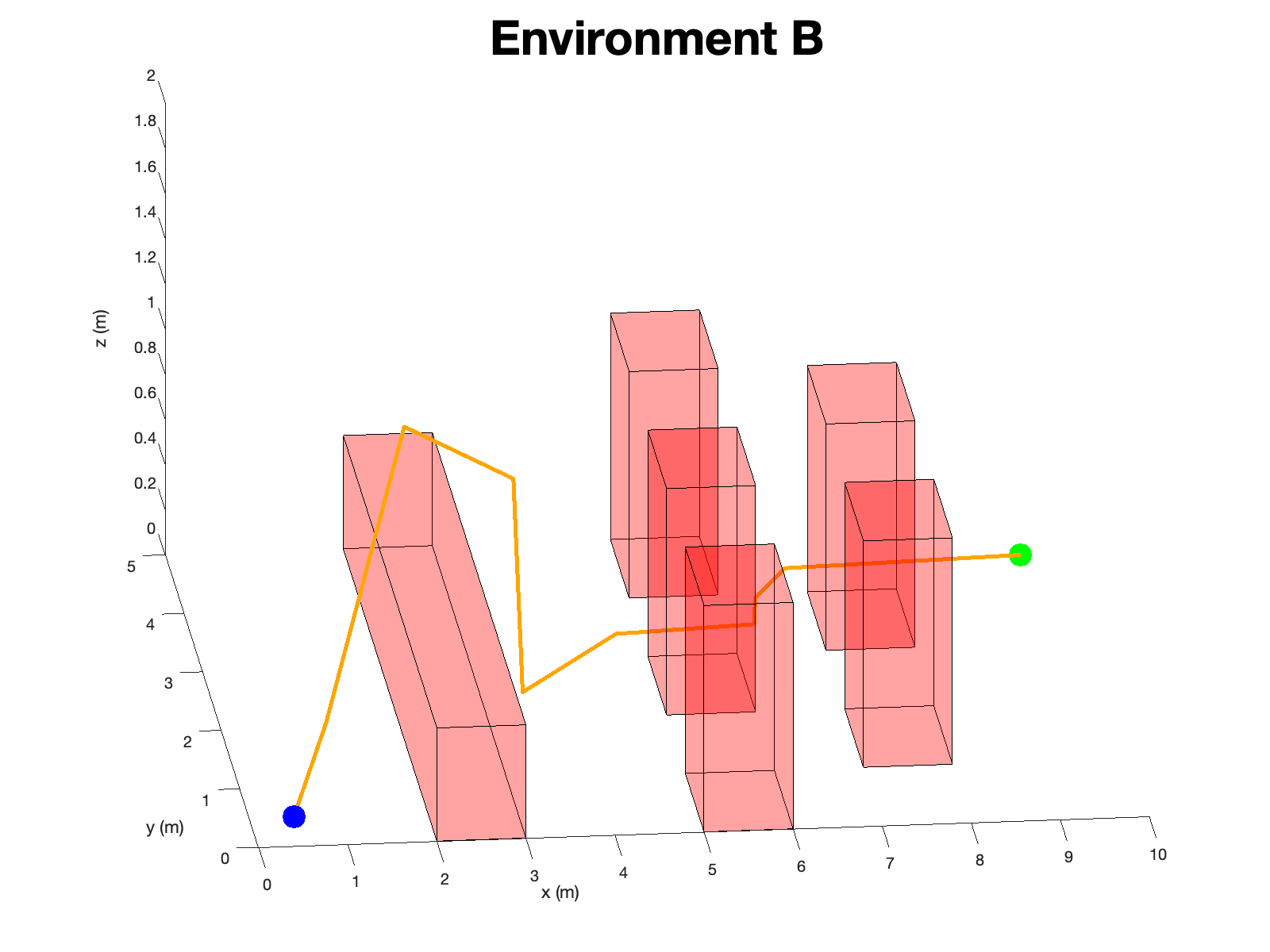}
    \includegraphics[width = 0.30\linewidth]{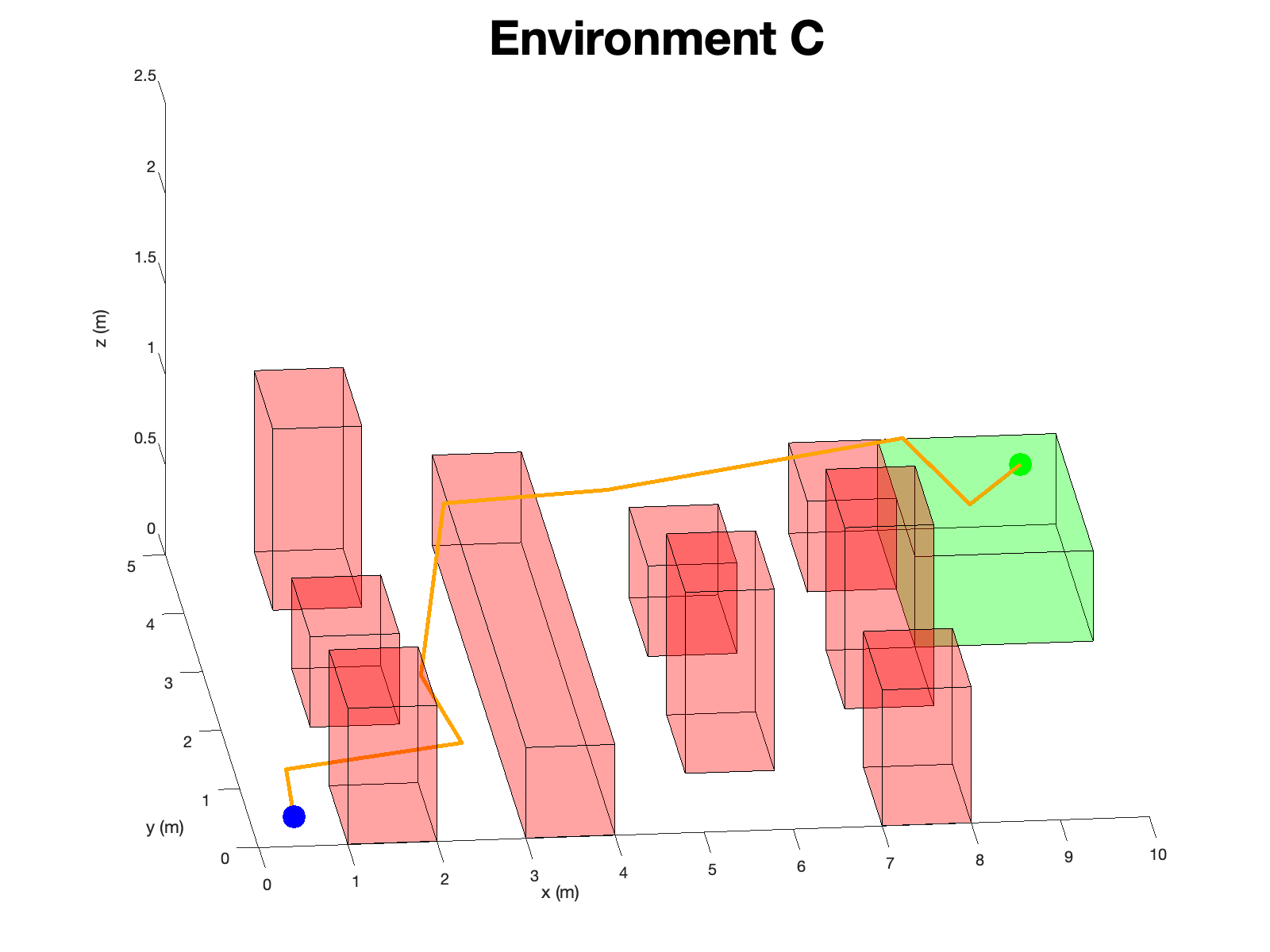}
    \caption{The trajectories generated by the path planning algorithm on three different environments. The environment A will be used in the Husky simulation for tracking the generated trajectory and show Husky's multi-locomotion capability.}
    \vspace{-0.1in}
    \label{fig:environments}
\end{figure*}

\begin{figure*}[t]
    \centering
    \vspace{0.1in}
    \includegraphics[width = \linewidth]{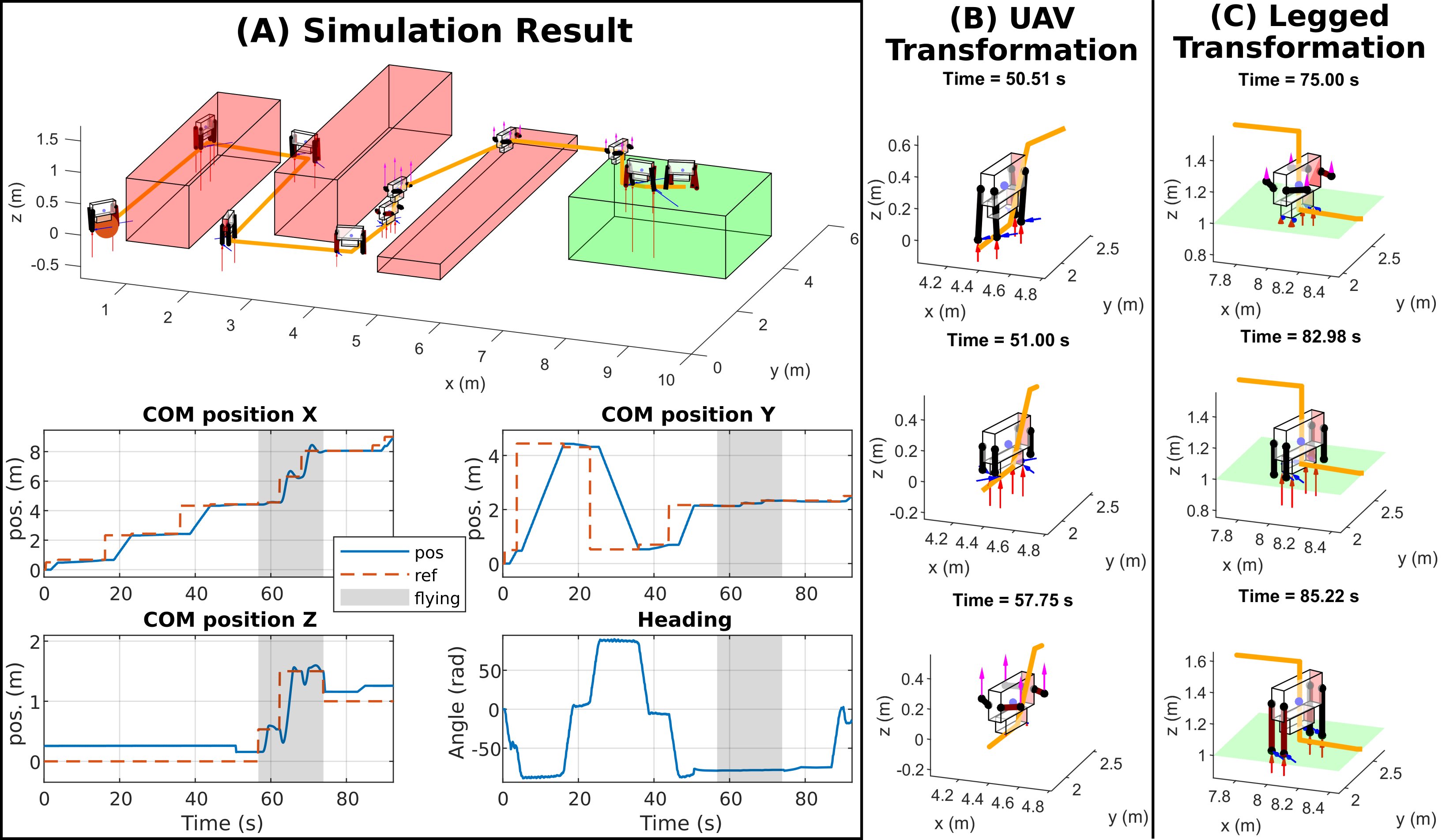}
    \caption{The simulation result for the trajectory following algorithm showing the legged and aerial mobility capabilities of Husky. \textbf{(A)} Shows the trajectory followed by the robot, position states, and heading in the simulation. \textbf{(B)} Shows the transformation sequence from legged to aerial mobility. \textbf{(C)} Shows the transformation sequence from aerial to legged mobility.}
    \vspace{-0.1in}
    \label{fig:results}
\end{figure*}

To calculate the locomotion cost including legged and aerial, it is necessary to not only determine the costs associated with each modes but also the cost corresponding to the transition from one mode to another. As such, the cost of transport on a walking edge denoted by $C_w$ is calculated using the power consumption at the joints $P_j$. Then, $P_j$ are integrated over the time of legged locomotion. The total joint power consumption is computed based on the torque and the angular velocity of each joint which is obtained from ROM. The time of legged locomotion is calculated based on the distance $d$ between the two nodes. As a result, $C_w$ is given by:
\begin{equation}
    C_w =  \int_0^{t_d} P_j(\tau)d\tau.
\end{equation}
%
The energetic cost on a flying edge $C_f$ is computed using the power consumption $P_f$ in hovering, the robot forward velocity $v_f$ in flying mode, and the altitude $z$ of the two nodes. Hence, $C_f$ is given by:
\begin{equation}
    C_f = P_f\frac{d}{v_f}+ mg(z_2 - z_1),
    \label{eq_Cf}
\end{equation}
\noindent where $z_1$ and $z_2$ are respectively the altitudes of the nodes 1 and 2, $m$ is the mass of Husky and $g$ the gravitational acceleration constant. Last, the transition cost $C_t$ between the two modes is determined based on the power consumption of the joints during the morphing process $P_t$. Then, $P_t$ is integrated over the time of transition $t_t$ which yields:
\begin{equation}
    C_t = \int_0^{t_t} P_s(\tau)d\tau .
\end{equation}
\noindent These three energetic costs are employed to determine the optimal path in the edge space generated by MM-PRM algorithm using the A$^\star$ algorithm.

\subsubsection{Find Optimal Path Using 3D A* Algorithm}

To find the optimal path in the graph, the A$^\star$ path search algorithm \cite{A_star} is used. The improved version of Dijkstra's algorithm \cite{dijkstra1959note} is employed to find the optimal path by using a heuristic function. The algorithm computes the best path to each node in order to only visit the most promising nodes. This avoids going through all possible paths and, therefore, finding the first-best optimal path with a low computational cost. Thus, each time the algorithm explores n-th node, it calculates the minimum cost $f(n)$ necessary to reach the goal by passing through it using the following formula:
\begin{equation}
    f(n) = g(n) + h(n),
\end{equation} 
\noindent where $g(n)$ is the real cost from the start to the n-th node, computed based on \eqref{eq:g(n)}, and $h(n)$ denotes the heuristic cost to the goal. The heuristic cost $h(n)$ is calculated by summing two conservative costs. First, the cost of walking on flat ground to the goal in a straight line is calculated. Second, the cost of flying vertically along the z-axis to the goal is obtained. Since the cost of walking is much lower than flying, this is the most optimal way to move between two points if there is no obstacles or impassable terrains. The following cost for $g(n)$ is defined:
\begin{equation}
    g(n) = \sum_{i = 0}^{E_w} C_{w,i} + \sum_{j = 0}^{E_f} C_{f,j} + N_t * C_{t}
    \label{eq:g(n)}
\end{equation}
\noindent where $E_w$ and $E_f$ are respectively the number of walking and flying edges traveled by Husky, $C_{w,i}$ is the cost on the walking edge $i$, $C_{f,j}$ denotes the cost on the flying edge $j$, and $N_t$ represents the number of transition made by Husky.

\section{Simulation and Result Discussions}
\label{sec:simulation}

\subsection{Environments and Path Planning Results}
\label{subsec:path_planning_results}

We designed several environments to test the path planning algorithm and the control architecture performance, as illustrated in Fig.~\ref{fig:environments}. We have placed box-shaped obstacles, and in some cases, the goal is located on a platform where the robot can walk. The purpose of these configurations is to put Husky in a situation where it has to perform at least one flight phase to reach the goal. Hence, it has to optimize its choice of locomotion mode to avoid obstacles and reduce its consumed energy. Fig.~\ref{fig:environments}, presents three of these environments and the planned path generated by our algorithm. 

We utilized the cost of transport of Husky as reported from our previous work \cite{ramezani_generative_2021}, and an estimation of energy consumption of the propeller motors during flight for a robot of this size and weight class. In the case of environment A shown in Fig.~\ref{fig:environments}, the cost of the optimized path is 9600 Joules while the direct one using only the flight mode is 14200 Joules. The use of the Husky's multi-modal locomotion, therefore, allows a very large gain (about 32\%) in terms of energy consumed, where most of the saving comes from the slower but much more energy efficient ground locomotion.


\subsection{Trajectory following simulation results}

We implement the waypoints generated in \ref{subsec:path_planning_results} for the robot to track and follow using the controller described in Fig.~\ref{fig:control_architecture}. In this simulation, we implemented the path generated for environment A, as shown in Fig.~\ref{fig:environments}. The robot was initialized on the ground and walks using a simple trotting gait as shown in the state machine described in Fig.~\ref{fig:control_architecture}, and fly using a simple flight controller to track the aerial trajectories. 

The simulation result can be seen in Fig.~\ref{fig:results}. The robot has successfully followed the desired trajectory and demonstrated the multi-modal locomotion capability that we proposed. Figure~\ref{fig:results} also shows the transformation sequence as the robot transition from the legged into the aerial mode and vise-versa. As shown in Fig.~\ref{fig:results}, the legged to aerial transformation can be achieved by crouching until the landing gear touches the ground, then the legs can safely reorient to the UAV configuration and starts flying. On the other hand, the aerial to legged transformation can be done in reverse: land, then reorient the legs to face the ground, and stand up to continue walking.

\section{Conclusions and Future Work}
\label{sec:conclusions}

In this paper, we presented the implementation of a high-level path planning algorithm based on MM-PRM and A$^\star$ algorithm to a legged-robot capable of both grounded and aerial movement. Both the high and low level control architecture are presented in this work and implemented in the simulator to show the multi-modal capabilities of our platform. The simulation has shown that the robot is capable of tracking the path found by the path-finding algorithm and is capable of transitioning from the legged to aerial mobility, and vice versa. In our future work, we will look into implementing the path finding algorithm in the lab environment and fully integrated the control architecture used in the simulation into the Husky Carbon for practical experiments.

\addtolength{\textheight}{-12cm}   

\printbibliography

@article{sihite_integrated_2021,
	title = {An {Integrated} {Mechanical} {Intelligence} and {Control} {Approach} {Towards} {Flight} {Control} of {Aerobat}},
	url = {http://arxiv.org/abs/2103.16566},
	urldate = {2021-07-05},
	journal = {arXiv:2103.16566 [cs, eess]},
	author = {Sihite, Eric and Darabi, Atefe and Dangol, Pravin and Lessieur, Andrew and Ramezani, Alireza},
	month = mar,
	year = {2021},
	note = {arXiv: 2103.16566},
}

@article{ramezani_generative_2021,
	title = {Generative {Design} of {NU}'s {Husky} {Carbon}, {A} {Morpho}-{Functional}, {Legged} {Robot}},
	url = {http://arxiv.org/abs/2104.05834},
	urldate = {2021-05-01},
	journal = {arXiv:2104.05834 [cs]},
	author = {Ramezani, Alireza and Dangol, Pravin and Sihite, Eric and Lessieur, Andrew and Kelly, Peter},
	month = apr,
	year = {2021},
	note = {arXiv: 2104.05834},
}

@article{ramezani_performance_2014,
	title = {Performance {Analysis} and {Feedback} {Control} of {ATRIAS}, {A} {Three}-{Dimensional} {Bipedal} {Robot}},
	volume = {136},
	issn = {0022-0434},
	doi = {10.1115/1.4025693},
	language = {en},
	number = {2},
	urldate = {2020-05-09},
	journal = {Journal of Dynamic Systems, Measurement, and Control},
	author = {Ramezani, Alireza and Hurst, Jonathan W. and Akbari Hamed, Kaveh and Grizzle, J. W.},
	month = mar,
	year = {2014},
	note = {Publisher: American Society of Mechanical Engineers Digital Collection},
}

@inproceedings{buss_preliminary_2014,
	address = {Chicago, IL, USA},
	title = {Preliminary walking experiments with underactuated {3D} bipedal robot {MARLO}},
	isbn = {978-1-4799-6934-0 978-1-4799-6931-9},
	doi = {10.1109/IROS.2014.6942907},
	language = {en},
	urldate = {2020-01-04},
	booktitle = {2014 {IEEE}/{RSJ} {International} {Conference} on {Intelligent} {Robots} and {Systems}},
	publisher = {IEEE},
	author = {Buss, Brian G. and Ramezani, Alireza and Akbari Hamed, Kaveh and Griffin, Brent A. and Galloway, Kevin S. and Grizzle, Jessy W.},
	month = sep,
	year = {2014},
	pages = {2529--2536},
}

@inproceedings{park_switching_2012,
	address = {St Paul, MN, USA},
	title = {Switching control design for accommodating large step-down disturbances in bipedal robot walking},
	isbn = {978-1-4673-1405-3 978-1-4673-1403-9 978-1-4673-1578-4 978-1-4673-1404-6},
	doi = {10.1109/ICRA.2012.6225056},
	language = {en},
	urldate = {2020-01-04},
	booktitle = {2012 {IEEE} {International} {Conference} on {Robotics} and {Automation}},
	publisher = {IEEE},
	author = {Park, Hae-Won and Sreenath, Koushil and Ramezani, Alireza and Grizzle, J.W.},
	month = may,
	year = {2012},
	pages = {45--50},
}

@article{park_finite-state_2013,
	title = {A {Finite}-{State} {Machine} for {Accommodating} {Unexpected} {Large} {Ground}-{Height} {Variations} in {Bipedal} {Robot} {Walking}},
	volume = {29},
	issn = {1552-3098, 1941-0468},
	doi = {10.1109/TRO.2012.2230992},
	language = {en},
	number = {2},
	urldate = {2020-01-04},
	journal = {IEEE Transactions on Robotics},
	author = {Park, Hae-Won and Ramezani, Alireza and Grizzle, J. W.},
	month = apr,
	year = {2013},
	pages = {331--345},
}

@incollection{ramezani_atrias_2012,
	title = {Atrias 2.0, a new 3d bipedal robotic walker and runner},
	isbn = {978-981-4415-94-1},
	urldate = {2020-05-09},
	booktitle = {Adaptive {Mobile} {Robotics}},
	publisher = {WORLD SCIENTIFIC},
	author = {Ramezani, Alireza and Grizzle, J.w.},
	month = may,
	year = {2012},
	doi = {10.1142/9789814415958_0060},
	pages = {467--474},
}

@article{dangol_towards_2020,
	title = {Towards thruster-assisted bipedal locomotion for enhanced efficiency and robustness},
	url = {http://arxiv.org/abs/2005.00347},
	urldate = {2020-05-09},
	journal = {arXiv:2005.00347 [cs, eess]},
	author = {Dangol, Pravin and Ramezani, Alireza},
	month = apr,
	year = {2020},
	note = {arXiv: 2005.00347},
}

@inproceedings{sharif_energy_2018,
	title = {Energy Efficient Path Planning of Hybrid Fly-Drive Robot ({HyFDR}) using A* Algorithm:},
	isbn = {978-989-758-321-6},
	url = {http://www.scitepress.org/DigitalLibrary/Link.aspx?doi=10.5220/0006912602010210},
	doi = {10.5220/0006912602010210},
	eventtitle = {15th International Conference on Informatics in Control, Automation and Robotics},
	pages = {201--210},
	booktitle = {Proceedings of the 15th International Conference on Informatics in Control, Automation and Robotics},
	publisher = {{SCITEPRESS} - Science and Technology Publications},
	author = {Sharif, Amir and Lahiru, H. M. and Herath, S. and Roth, Hubert},
	urldate = {2022-03-30},
	date = {2018},
}

@article{dijkstra1959note,
  title={A note on two problems in connexion with graphs},
  author={Dijkstra, Edsger W and others},
  journal={Numerische mathematik},
  volume={1},
  number={1},
  pages={269--271},
  year={1959}
}

@inproceedings{sharif2019new,
  title={A new algorithm for autonomous outdoor navigation of robots that can fly and drive},
  author={Sharif, Amir and Choi, Sunghoon and Roth, Hubert},
  booktitle={Proceedings of the 5th International Conference on Mechatronics and Robotics Engineering},
  pages={141--145},
  year={2019}
}

@ARTICLE{MM_PRM,
  author={Kavraki, L.E. and Svestka, P. and Latombe, J.-C. and Overmars, M.H.},
  journal={IEEE Transactions on Robotics and Automation}, 
  title={Probabilistic roadmaps for path planning in high-dimensional configuration spaces}, 
  year={1996},
  volume={12},
  number={4},
  pages={566-580},
  doi={10.1109/70.508439}
}

@inproceedings{araki_multi-robot_2017,
	title = {Multi-robot path planning for a swarm of robots that can both fly and drive},
	doi = {10.1109/ICRA.2017.7989657},
	eventtitle = {2017 {IEEE} International Conference on Robotics and Automation ({ICRA})},
	pages = {5575--5582},
	booktitle = {2017 {IEEE} International Conference on Robotics and Automation ({ICRA})},
	author = {Araki, Brandon and Strang, John and Pohorecky, Sarah and Qiu, Celine and Naegeli, Tobias and Rus, Daniela},
	date = {2017-05},
}

@article{suh_optimal_2019,
	title = {Optimal Motion Planning for Multi-Modal Hybrid Locomotion},
	author = {Suh, Hyung Ju and Xiong, Xiaobin and Singletary, Andrew and Ames, Aaron and Burdick, Joel},
	date = {2019-09-23},
}

@ARTICLE{A_star,  
    author={Hart, Peter E. and Nilsson, Nils J. and Raphael, Bertram},  journal={IEEE Transactions on Systems Science and Cybernetics},   title={A Formal Basis for the Heuristic Determination of Minimum Cost Paths},   
    year={1968},  
    volume={4},  
    number={2},  
    pages={100-107},  
    doi={10.1109/TSSC.1968.300136}
}

@article{raibert1984experiments,
  title={Experiments in balance with a 3D one-legged hopping machine},
  author={Raibert, Marc H and Brown Jr, H Benjamin and Chepponis, Michael},
  journal={The International Journal of Robotics Research},
  volume={3},
  number={2},
  pages={75--92},
  year={1984},
  publisher={Sage Publications Sage CA: Thousand Oaks, CA}
}

@article{raibert2008bigdog,
  title={Bigdog, the rough-terrain quadruped robot},
  author={Raibert, Marc and Blankespoor, Kevin and Nelson, Gabriel and Playter, Rob},
  journal={IFAC Proceedings Volumes},
  volume={41},
  number={2},
  pages={10822--10825},
  year={2008},
  publisher={Elsevier}
}

@inproceedings{liang2021rough,
  title={Rough-Terrain Locomotion and Unilateral Contact Force Regulations With a Multi-Modal Legged Robot},
  author={Liang, Kaier and Sihite, Eric and Dangol, Pravin and Lessieur, Andrew and Ramezani, Alireza},
  booktitle={2021 American Control Conference (ACC)},
  pages={1762--1769},
  year={2021},
  organization={IEEE}
}

@article{sihite2021optimization,
  title={Optimization-free Ground Contact Force Constraint Satisfaction in Quadrupedal Locomotion},
  author={Sihite, Eric and Dangol, Pravin and Ramezani, Alireza},
  journal={arXiv preprint arXiv:2111.12557},
  year={2021}
}

@inproceedings{sihite2021unilateral,
  title={Unilateral Ground Contact Force Regulations in Thruster-Assisted Legged Locomotion},
  author={Sihite, Eric and Dangol, Pravin and Ramezani, Alireza},
  booktitle={2021 IEEE/ASME International Conference on Advanced Intelligent Mechatronics (AIM)},
  pages={389--395},
  year={2021},
  organization={IEEE}
}

@inproceedings{dangol2020performance,
  title={Performance satisfaction in midget, a thruster-assisted bipedal robot},
  author={Dangol, Pravin and Ramezani, Alireza and Jalili, Nader},
  booktitle={2020 American Control Conference (ACC)},
  pages={3217--3223},
  year={2020},
  organization={IEEE}
}

@inproceedings{dangol2021hzd,
  title={A HZD-based Framework for the Real-time, Optimization-free Enforcement of Gait Feasibility Constraints},
  author={Dangol, Pravin and Lessieur, Andrew and Sihite, Eric and Ramezani, Alireza},
  booktitle={2020 IEEE-RAS 20th International Conference on Humanoid Robots (Humanoids)},
  pages={156--162},
  year={2021},
  organization={IEEE}
}

@inproceedings{de2020thruster,
  title={Thruster-assisted center manifold shaping in bipedal legged locomotion},
  author={de Oliveira, Arthur CB and Ramezani, Alireza},
  booktitle={2020 IEEE/ASME International Conference on Advanced Intelligent Mechatronics (AIM)},
  pages={508--513},
  year={2020},
  organization={IEEE}
}

\end{document}